%% file: main.tex
\documentclass[10pt,twocolumn,letterpaper]{article}

\usepackage{cvpr}
\usepackage{times}
\usepackage{epsfig}
\usepackage{graphicx}
\usepackage{amsmath}
\usepackage{amssymb}
\usepackage[dvipsnames]{xcolor}
\usepackage{subcaption}
\usepackage{comment}
\usepackage{enumitem}
\usepackage{float} 
\usepackage{graphicx}

\include{headers/tim}

\usepackage[pagebackref=true,breaklinks=true,letterpaper=true,colorlinks,bookmarks=false]{hyperref}

\def\cvprPaperID{6323} 

\cvprfinalcopy

\ifcvprfinal\fi
\begin{document}

\title{Towards Robust Deep Neural Networks}

\author{Timothy E. Wang\\
United Technologies  \\
Research Center \\
{\tt\small wangte@utrc.utc.com }
\and
Yiming Gu\thanks{Work carried out while at UTRC}\\
Uber\\
{\tt\small yiming.guu@gmail.com}
\and 
Dhagash Mehta\\
United Technologies  \\
Research Center \\
{\tt\small dhagashbmehta@gmail.com}
\and 
Xiaojun Zhao\\
United Technologies  \\
Research Center \\
{\tt\small zhaox@utc.utrc.com}
\and 
Edgar A. Bernal\thanks{Work carried out while at UTRC}\\
University of Rochester\\
{\tt\small edgar.bernal@rochester.edu}
}

\maketitle
\begin{abstract}
\input{abstract}
\end{abstract}

\section{Introduction}
\label{sec:intro} 
\input{01intro}

\section{Background and Related Work}
\label{sec:related}
\input{01related}

\section{Measure of Robustness via Sensitivity Analysis} 
\label{sec:sense} 
\input{02sense}

\section{Optimization Landscape Minima: Energy Value vs. Sensitivity} 
\label{sec:energy} 
\input{03energy}

\section{Learning with Sensitivity Minimization} 
\label{sec:training} 
\input{05training}

\section{Experimental Results} 
\label{sec:results} 
\input{06results}

\section{Conclusion and Future Work} 
\input{07conclu}

\bibliographystyle{abbrv}
\bibliography{refs/tim,refs/daghesh,refs/edgar}

\appendix
\clearpage
\onecolumn
\section{Appendix - Section 6: Experimental Results}
The plots included in this appendix illustrate the evolution of the different losses through the training process of the methods being compared. 
\subsection{Experiments on Sensitivity-Based Optimization (Sec. 6.1)}
\input{appendix02} 
\subsection{Experiments on Adversarial-Margin-Based Optimization (Sec. 6.2)}
\input{appendix01}

\end{document}

%% file: headers/tim.tex
\newcommand{\benum}{\begin{enumerate}}
\newcommand{\eenum}{\end{enumerate}}
\newcommand{\nc}{\newcommand}
\newcommand{\rnc}{\renewcommand}
\newcommand{\bvb}{\begin{verbatim}}
\newcommand{\evb}{\end{verbatim}}

\nc{\<}{\langle}
\rnc{\>}{\rangle}
\nc{\bq}{\textbf}
\nc{\m}{\textrm}
\nc{\bb}{\mathbb}
\nc{\til}{\texttildelow}
\nc{\dps}{\displaystyle}
\rnc{\l}{\left(}\rnc{\r}{\right)}
\nc{\lc}{\left\{}\nc{\rc}{\right\}}
\nc{\lb}{\left[}\nc{\rb}{\right]}
\nc{\ba}[1]{\begin{array}{#1}}
\nc{\ea}{\end{array}}       
\nc{\ra}{\rightarrow}
\nc{\li}{\left |}
\nc{\ri}{\right |}
\nc{\pde}[2]{\frac{\partial #1}{\partial #2}}
\nc{\ode}[2]{\frac{d #1}{d #2}}
\nc{\odee}[3]{\frac{d^{#3} #1}{d #2^{#3}}}
\nc{\pdee}[3]{\frac{\partial^{#3} #1}{\partial #2^{#3}}}
\nc{\bn}{\begin{enumerate}}
\nc{\en}{\end{enumerate}}
\nc{\bt}{\begin{theorem}}
\nc{\et}{\end{theorem}}
\nc{\y}[1]{\lambda_{#1}}
\nc{\ninf}{{\oplus}^{-\infty}}
\nc{\pinf}{{\oplus}^{+\infty}}
\nc{\nninf}{{\otimes}^{-\infty}}
\nc{\ppinf}{{\otimes}^{+\infty}}
\nc{\ir}{\mathbb{I}\mathbb{R}}
\nc{\closure}[2][3]{{}\mkern#1mu\overline{\mkern-#1mu#2}} 
\nc{\ep}{\mathcal{E}_{P}}
\nc{\mr}{\mathcal{M}_{r}}
\nc{\mfa}{\mathcal{M}_{f,a}}
\nc{\mfp}{\mathcal{M}_{f,p}}
\nc{\mt}{\m{T}}
\nc{\dB}{\textit{dB}}
\nc{\VM}{\delta} 
\nc{\GM}{\Theta}
\nc{\PM}{\phi} 
\nc{\x}{\mathbf{x}}

\usepackage{xcolor}

\newcommand{\R}{\mathbb{R}}

\newtheorem{theorem}{Theorem}

%% file: abstract.tex
We investigate the topics of sensitivity and robustness in feedforward and convolutional neural networks. Combining energy landscape techniques developed in computational chemistry with tools drawn from formal methods, 
we produce empirical evidence indicating that networks corresponding to lower-lying minima in the optimization landscape of the learning objective tend to be more robust. 
The robustness estimate used is the inverse of a proposed sensitivity measure, which we define as the volume of an over-approximation of the reachable set of network outputs under all additive $l_{\infty}$-bounded perturbations on the input data. We present a novel loss function which includes a sensitivity term in addition to the traditional task-oriented and regularization terms. In our experiments on standard machine learning and computer vision datasets, we show that the proposed loss function leads to networks which reliably optimize the robustness measure as well as other related metrics of adversarial robustness without significant degradation in the classification error. Experimental results indicate that the proposed method outperforms state-of-the-art sensitivity-based learning approaches with regards to robustness to adversarial attacks. We also show that although the introduced framework does not explicitly enforce an adversarial loss, it achieves competitive overall performance relative to methods that do.

%% file: 01intro.tex
The advent of machine learning techniques, 
most notably deep learning \cite{bengio2015deep}, has made it possible to automate
complex tasks such as vision-based object detection, 
natural language processing, machine translation, and stock-market analysis. 
Despite its tremendous success in many academic endeavors and commercial applications, the adoption of deep 
learning in the perception, decision and control loops of 
mission/safety-critical systems has been limited. 
One possible reason for the slow rate of adoption is the limited 
theoretical understanding of the inner workings of deep neural networks (DNNs). 
Another possible reason is the lack of guarantees of certain behavioral, and, in particular, robustness properties. 
One such robustness property is the ability of the network to be resistant to adversarial attacks. 
In an adversarial attack, the input data is perturbed minimally 
such that, while the resulting adversarial example closely resembles the unmolested sample, the output of the trained network is affected. 
Recent work in the deep learning literature
~\cite{goodfellow6572explaining,papernot2016cleverhans,szegedy2013intriguing, 
kurakin2016adversarial,huang2017adversarial,song2017inaudible,evtimov2017robust}
has shown that DNNs may be susceptible to adversarial examples. For state-of-art
DNN image classifiers, researchers have concocted methods to reliably engineer small perturbations that 
result in successful adversarial examples. In one such example~\cite{goodfellow6572explaining}, the generated adversarial example looks, to the naked human eye, indistinguishable from the original sample, but produces a drastically different classifier output. Some input perturbations
are synthetic~\cite{goodfellow6572explaining}, while others target physical, real-world objects~\cite{kurakin2016adversarial,evtimov2017robust}. 

In traditional verification and validation (V\&V) processes for safety-critical systems,  
the robustness of the system is assured via an extensive testing procedure which
samples the expected variations in the values of the parameters 
of the operating environment and system inputs until certain criteria of appropriate coverage metrics are reached. 
This test-based approach is likely to be less effective for systems with deep learning components since
the dimensions of their input spaces are typically much larger. 
Furthermore, the lack of understanding of the topology of the input spaces involved leads to 
an open theoretical question: how do we define covergence metrics for V\&V tests applied to systems with deep learning components?  
Nevertheless, we know that it is feasible to 
provide formal guarantees of robustness properties of small to medimum DNNs~\cite{Katz2017} as well as to enforce the guarantee of a robustness property through the training process for larger DNNs~\cite{Kolter2017}. 
With the increase in predictability that comes with the 
formal guarantees of adversarial robustness, it is conceivable that deep learning systems could
be deployed in mission and/or safety-critical systems. 

In this paper, we present a new learning framework that results in networks which are less sensitive to changes in the input, and, consequently, likely to be more robust to adversarial attacks. The method reliably yields networks of increased robustness by introducing a cost term that penalizes output sensitivity to changes in the input. 

\begin{figure}
\includegraphics[width=0.45\textwidth]{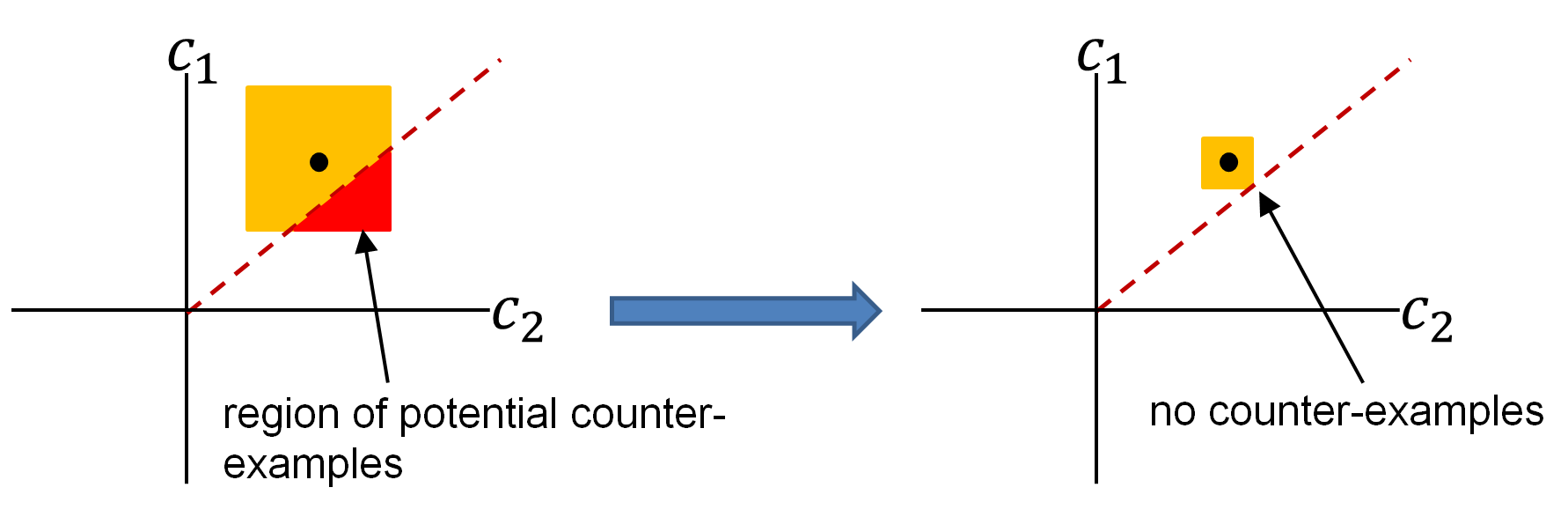}
\caption{Reducing the size of the output reachable set could lead to increased robustness to adversarial inputs}
\label{fig:hyp01} 
\end{figure}

The sensitivity measure is described briefly below and is further detailed in Sec.~\ref{sec:sense}.  Consider a portion of a two-dimensional output space illustrated in Fig.~\ref{fig:hyp01} (left) where a segment of the decision boundary between classes  $C_1$ and $C_2$ lies.  The black dot represents the network output for a sample $x$ belonging to $C_1$.  Consider a set $\Delta$ containing all possible perturbation vectors expected upon network deployment.  These could include, in the case of a vision-based object classification application, synthetic noise, physical perturbations, shadows, glare, etc.    
Consider the input data $x$, and the set $\mathcal{X}$ resulting from perturbing $x$ with all possible disturbances from $\Delta$. 
The DNN maps the perturbed input set $\mathcal{X}$
into some output reachable set, an over-approximation of which is represented by the box in the figure. 
We measure the \textit{sensitivity} of the network as the aggregate volume of all over-approximations of the output reachable set across the training samples for a predetermined $\Delta$. 

Also shown in Fig.~\ref{fig:hyp01} (left) is the region of potential
counter-examples (red portion of the box), which contains all the points in the over-approximated output set for which the classification changes from $C_1$ to $C_2$,
i.e., potential outputs of adversarial inputs. 
We hypothesize that, by reducing the volume of the output reachable set for some given perturbed input set, the
possibility of successful adversarial inputs can be ameliorated, as indicated by the smaller orange box in Fig.~\ref{fig:hyp01} (right). 
 
The main contributions of this paper can be summarized as follows: \begin{itemize}[leftmargin=*, itemsep=0pt]
  \item We propose a novel sensitivity metric for feedforward neural networs and introduce two different approaches to estimate it (see Sec.~\ref{sec:sense}): one based on Reluplex~\cite{Katz2017}, an update on the classical Simplex method that enables support of ReLU constraints; and one based on evaluating the dual cost of a Linear Programming (LP) relaxation of the exact encoding of the input-output mapping of a ReLU-based network~\cite{Kolter2017,Krish2018}. 
  \item We conduct an empirical study on the relationship between the sensitivities of neural networks and the energy values of their corresponding optimization landscape minima, from which we conclude that networks corresponding to lower-lying minima tend to be less sensitive than those corresponding to higher-valued minima.  We also show that traditional optimization approaches based on Stochastic Gradient Descent (SGD) algorithms and regularization do not necessarily lead to networks with good sensitivity properties (see Sec.~\ref{sec:energy}).
  \item Based on the above results, we propose a novel loss function which enables effective task-oriented learning while penalizing high network sensitivity to changes in the input (see Sec.~\ref{sec:training}).  We empirically verify the effectiveness of the proposed cost function, and compare the performance of our method with that of state-of-art approaches (see Sec.~\ref{sec:results}).  
\end{itemize}
%
%
%

%% file: 01related.tex
%
%

Many supervised machine and deep learning algorithms operate based on the principle of Empirical Risk Minimization (ERM), whereby the expectation of a loss function associated with a given hypothesis is approximated with its empirical estimate.  The main implication of this modus operandi is the so-called Generalization Error (GE), which refers to the difference between the empirical error and the expected error; practically speaking, the GE manifests itself in a difference in algorithm performance on unseen data relative to the performance observed on the training data.  Since ERM is an optimistically biased estimation process \cite{RandG}, techniques to bridge (and understand) the gap between the empirical risk and the true risk have been proposed: the narrower the gap, the better the generalization properties of the algorithm.  Broadly speaking, these techniques fall under one of the following categories:
\subsection{Sensitivity and Robustness Analysis and Optimization}
The authors of \cite{RandG} define robustness as the property of a machine learning algorithm to produce similar test and training errors when the test and training samples are similar to each other, and hypothesize that a weak measure of robustness is both sufficient and necessary for good generalizability. Furthermore, they derive a generalization bound that holds under certain constraints.  The authors of \cite{novak2018sensitivity} conclude empirically that trained neural networks are more robust to input perturbations in the vicinity of the training manifold and that higher robustness correlates well with good generalization properties.  In \cite{Sokolic2017RobustLM}, the relation between the GE and the classification margin of a neural network is studied.  The authors conclude that a necessary condition for good generalization is for the Jacobian matrix of the network to have a bounded spectral norm in the neighborhood of the training samples, in line with the results presented in \cite{novak2018sensitivity,Yoshida2017SpectralNR,DBLP:journals/corr/abs-1803-08680}.
\subsection{Robustness Against Adversarial Attacks}
Adversarial examples are data samples constructed by applying perturbations to known training 
samples along directions that result in the largest change at the network output \cite{Goodfellow2014ExplainingAH}, so that the perturbed input, often indistinguishable from the original input, leads to an erroneous decision by the network. 
This vulnerability is directly related to the GE and can be ameliorated by 
training with adversarial examples \cite{Goodfellow2014ExplainingAH}.  
The approach introduced in \cite{Gu2014TowardsDN} aims at achieving robustness against adversarial attacks, 
and consists of stacking a denoising auto encoder (DAE) with the network of interest, 
the role of the DAE being to map back the adversarial example to the training manifold.  
An alternative architecture based on contractive auto encoders (CAE) is also proposed. 
The authors of~\cite{Kolter2017} 
propose a method to train deep networks based on ReLU activation units 
that are provably robust to adversarial attacks; they achieve this by minimizing the worst-case loss across the
a convex over-approximation of the reachable set under norm-bounded input perturbations.
\subsection{Formal Verification of Neural Networks}
Often, the GE includes behavior that is not only incorrect, but also unpredictable, 
which prevents deployment of deep learning algorithms in safety-critical applications.
Formal verification refers to techniques aimed at providing mathematical guarantees 
about the behavior of systems including computer programs~\cite{Peled}. 
Formal verification is used most often for 
safety-critical applications in sectors such as aerospace, nuclear and
rail.  
Early approaches to formally verifying neural networks leveraged SMT solvers~\cite{Pulina2012ChallengingSS}, 
but did not scale well and were practical on very small networks with 
a single hidden layer and a small number of neurons.  
Reluplex~\cite{Katz2017} extended the simplex algorithm to support 
the piece-wise linear nature of ReLU activation units, 
and showed improved scalability properties.  
More recently, authors have tackled the problem using
optimization-based approaches including: Mixed Integer Programming (MIP) 
formulations~\cite{cheng2017maximum} with local search~\cite{dutta2018output}, 
LP relaxation~\cite{Ehlers2017}, 
branch and bound~\cite{Ehlers2017,bunel2017piecewise}, and dual formulations~\cite{Krish2018, Kolter2017}. 
In contrast to most of the other literature, the framework 
introduced in~\cite{Krish2018} applies to a general class of activation functions.
\subsection{Analysis of Optimization Landscapes}
The training of modern machine learning models, most notably of deep networks, is posed in the form of highly non-convex optimization tasks with multiple local minima. The simplest way to quantify the quality of landscape minima is to measure their distance to the global minimum: the closer the energy of a local minimum to that of the global minimum, the better the quality of the minimum and its corresponding network. This definition is ill-posed as in general the global minimum is not known for a typical nonlinear optimization function. One of the dominant explanations as to why deep learning works so well is that there may be no bad minima at all for the loss functions of deep network \cite{baldi1989neural,saxe2013exact,NguyenLoss,goodfellow2014qualitatively,soudry2016no,choromanska2014loss}: under certain unrealistic assumptions, it has been shown that all the local minima are global minima \cite{kawaguchi2016deep}. In more realistic scenarios, cost functions with a non-vanishing $l_2$-regularization term have been shown to have local minima which are not global minima \cite{Mehta2018:deeplinear,NIPS2017_6844}. Another branch of research leverages optimization methods developed for chemical physics, called the potential energy landscape theory \cite{Wales04}. Using such techniques, the authors of \cite{BallardDMMSSW17} showed that the landscape of the loss function for a specific feed-forward artificial neural network with one hidden layer trained on the MNIST dataset exhibited a single funnel structure. In a recent paper \cite{PhysRevE.97.052307} (see also, \cite{2017arXiv170610239W,kawaguchi2017generalization}), the measure of the goodness of a minimum is refined by adding further metrics in addition to the value of the cost function or the performance. The work concluded that when an overspecified artificial neural network with one hidden layer and $l_2$-regularization was used to learn the exclusive OR (XOR) function, although the classification error was often zero for various minima, the sparsity structure of the network varied with the minima.

%% file: 02sense.tex
\begin{figure}
\centering
\includegraphics[width=0.25\textwidth]{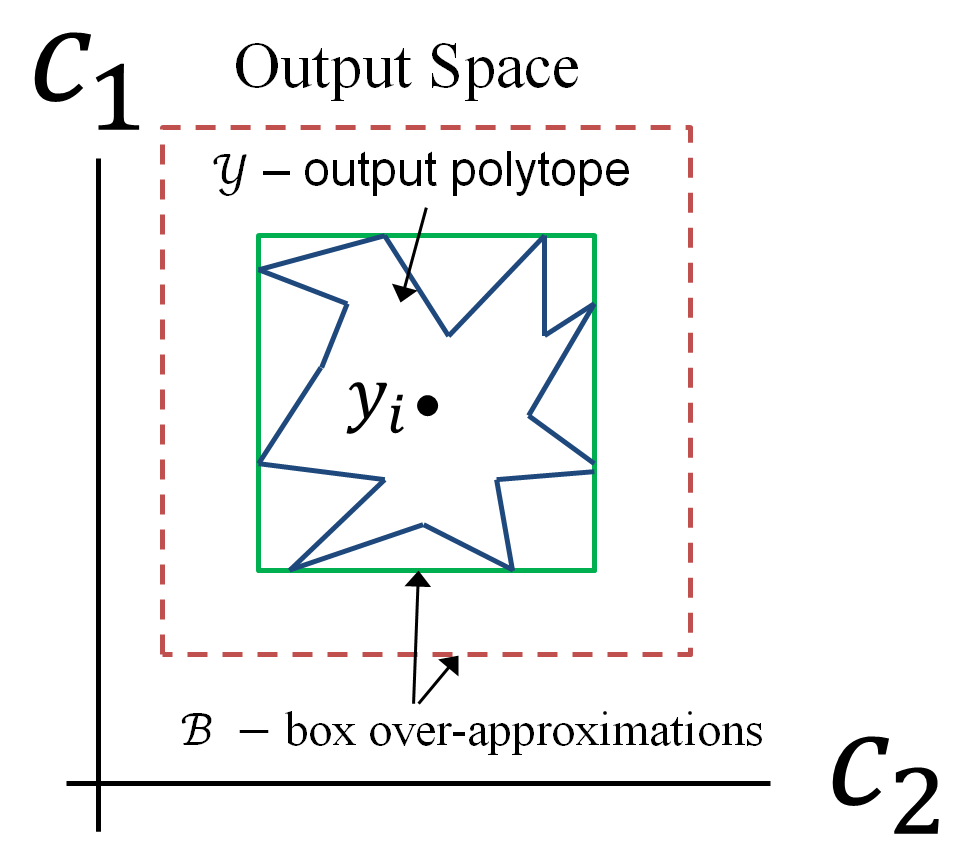} 
\caption{Proposed sensitivity measure is the volume of the box over-approximation of the output polytope which results from $l_{\infty}$-norm bounded perturbations to the input.}
\label{fig:over} 
\end{figure}

This section provides a detailed description of the sensitivty measure overviewed in Sec.~\ref{sec:intro}, as well as
of the methods that we use to estimate sensitivity for feedforward neural networks with ReLU activation nodes under 
$l_{\infty}$-norm bounded perturbations. Throughout the paper, we use ReLU as the activation function.
A feedforward neural network is a parameterized function $f_{W}$ which maps
input data $x_i$ in $\R^{n}$ to output vectors $y_i$ in $\R^{m}$. 
By applying $l_{\infty}$-norm bounded perturbations to an input data vector $x_i$, we obtain a
perturbed input set $\mathcal{X}=\lc x |x\in\R^{n}, \|x - x_i\|_{\infty} \leq \epsilon\rc$, where $\epsilon$ denotes the perturbation bound.
Given the perturbed input set $\mathcal{X}$, the output reachable set of the network 
is $\mathcal{Y}=\lc y \in \R^{m} | y=f_{W}(x) \wedge x\in\mathcal{X}\rc$. As stated earlier, we use an estimate of the volume of $\mathcal{Y}$ as a surrogate metric for network sensitivity.
For feedforward networks with ReLU activation nodes, $\mathcal{Y}$ is generally a non-convex polytope
whose exact volume is difficult to compute. 
Instead, we define the sensitivity of the network as the volume of a box 
over-approximation of the output reachable set as illustrated in Fig.~\ref{fig:over}.  
The over-approximation can be either tight (green box with solid boundary in Fig.~\ref{fig:over}) 
or loose (red box with dashed boundary in Fig.~\ref{fig:over}) depending 
on the computational methods used.  There is a significant 
trade-off between the conservatism of the over-approximation and the time 
it takes to compute it. 
We now describe the two methods used in our experiments. 
The first method, based on the dual formulation, 
is used to compute a conservative (i.e., loose) over-approximation. 
This method turned out to be efficient enough in practice to be incorporated in the training process 
described in Sec.~\ref{sec:training}. 
The second method uses Reluplex, a SMT solver specialized for DNNs, to compute a tight
over-approximation of the output polytope. 

\subsection{Computing Sensitivity Using a Dual Formulation} 
\label{subsec:dual}
The volume of the tight over-approximation (green box in Fig.~\ref{fig:over}) 
can be computed by finding the min and max values along each dimension of the output vector and then 
computing the product of the lengths of the resulting intervals across all dimensions. 
Finding the min and max of the output dimensions of a $N$-layer ReLU network requires solving 
a set of difficult optimization problems i.e., minimize and maximize $y[i], i=1,\ldots,m$, where 
$y[i]$ is the $i$-th entry of the output vector $y$, 
subject to a set of piece-wise linear constraints imposed by the ReLU activations. 
The min of $y[i]$ (denoted $y_{min}[i]$) under $l_{\infty}$-norm bounded perturbations to input $x$ with bound $\epsilon$ is the solution of the following set of optimization problems: 
\begin{align}
	& \underset{}{\text{minimize}} & 
	& y[i],\quad i=1,\ldots,m \label{eq:opt01}\\
    &  \text{subject to} 
	& &  x[j]-\epsilon \leq \hat{x}[j] \leq x[j] + \epsilon,\quad j=1,\ldots,n \label{eq:box} \\
& & & z_1 = W_0 \hat{x} + b_0  \label{eq:linear}  \\
 & & & y = W_{N} z_{N} + b_N  \nonumber \\
 & & & z_{k+1} = \max{\left( 0, W_k z_k + b_k\right )},\quad k=1,\ldots,N-1 \label{eq:pwlinear} 
 \end{align}
where $N$ is the number of layers in the network, and $W_k$ and $b_k$ are the weights and biases between layers $k$ and $k+1$, respectively.
The constraints in Eq.~\ref{eq:box} capture the fact that the input belongs to the perturbed set $\mathcal{X}$. 
The piece-wise linear constraints in Eq.~\ref{eq:pwlinear} denote the relations between the 
inputs and outputs of the layers with ReLU activations. 
The max of $y[i]$ (denoted $y_{max}[i]$) is just the negative of the min of $-y[i]$ in Eq.~\ref{eq:opt01}. 
The volume of the over-approximation is: 
\begin{equation} 
	\Pi_{i=1}^{m} y_{max}[i] - y_{min}[i]
\label{eq:volume}
\end{equation} 
It is known within the operational research community that the optimization problem in~\ref{eq:opt01} can be transformed into a
mixed-integer linear programming (MILP) problem through the usage of the big-$M$ trick~\cite{Richards2005}. 
Therefore computing the sensitivity of a network can be done by solving a MILP 
using a state-of-art solver such as Gurobi~\cite{Optimization2014} or 
by using other techniques with exact encodings benchmarked in recent work~\cite{bunel2017piecewise}. 
However, it remains difficult to scale up the ``exact" methods to efficiently verify sensitivity of
networks with multiple fully-connected layers and more than hundreds of ReLU nodes.  

\begin{figure}
\centering
\includegraphics[width=0.25\textwidth]{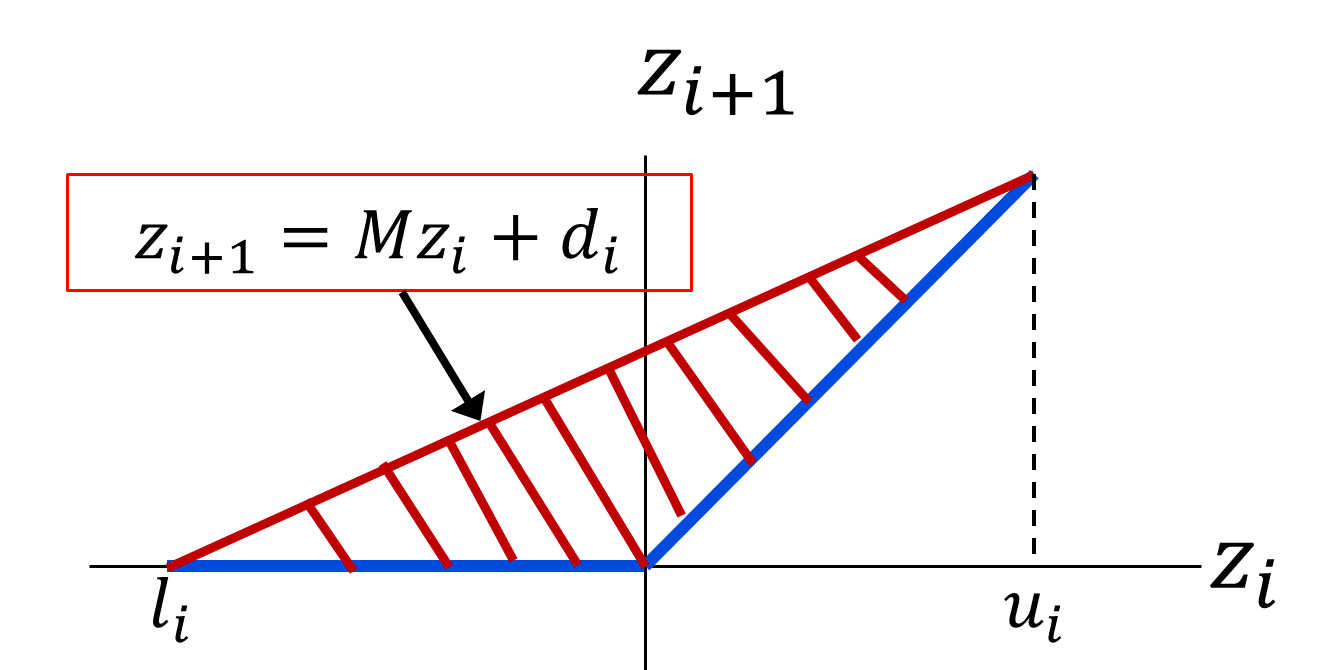} 
\caption{Linear relaxation of piece-wise linear constraints.} 
\label{fig:linear} 
\end{figure}

Due to the associated computational complexity, using methods with exact encodings is not yet practical for 
our main goal, which is to incorporate the sensitivity measure into the loss function driving the learning. Instead, we adopt an approach inspired by recent work in which ReLU constraints are relaxed into a set of linear constraints. The relaxation is illustrated graphically in Fig.~\ref{fig:linear}. 
Consider the piece-wise linear constraints $z_{i+1} = \max{\left( 0, W_i z_i + b_i\right )} $
from Eq.~\ref{eq:pwlinear}; under the relaxation procedure, they are transformed into a set of linear constraints:  
\begin{equation} 
	\begin{array}{l} 
		z_{i+1} \leq M z_{i} + d_i \\  
		l_i \leq z_{i} \leq u_i \\ 
		z_{i+1}\geq z_{i}\\ 
		z_{i+1} \geq 0 \\
	\end{array}
	\label{eq:relaxed} 
\end{equation} 
With the linear constraints in Eq.~\ref{eq:relaxed}, the 
optimization task in Eq.~\ref{eq:opt01} is transformed into a
primal linear programming (LP) problem.  
Furthermore, the dual objective, when evaluated at any feasible
point, becomes a lower-bound to the primal objective. 
As noted in~\cite{Kolter2017}, there is a feasible point which 
can guarantee in practice a ``sufficiently tight enough" lower bound to the primal objective. 

Let $J(x,C,\epsilon,\nu; W)$ be the dual objective function from~\cite{Kolter2017}, where $x$ is the input data, $C$ is the primal objective matrix, 
$\epsilon$ is the $l_{\infty}$ norm bound of the allowable input perturbations, $\nu$
is the dual variable, and $W$ represents the network parameters; the \textit{sensitivity function} $S(x,\epsilon, \nu; W)$, which measures the sensitivity of a neural network under norm-bounded perturbations, is given by:
\begin{equation} 
\begin{aligned}
	S(x,\epsilon, \nu; W) & = & \prod_{j=1}^{m} -J(x,-I_{m\times m},\epsilon,\nu; W)[j] \\
			      &  & -J(x,I_{m\times m} , \epsilon, \nu; W)[j]
\end{aligned} 
\label{eq:sense} 
\end{equation} 
where $I_{m\times m} \in \mathbb{R}^{m\times m}$ is the identity matrix.  Eq.~\ref{eq:sense} is the volume of a (box) over-approximation of the output reachable set of a feed-forward neural network with ReLU activation nodes under 
$l_{\infty}$-norm bounded perturbations with bound $\epsilon$. 

The sensitivity function can be evaluated very efficiently on a GPU-enabled laptop with CUDA for networks with multiple layers, and up to a few thousand nodes. Comparatively, the sensitivity of a similar-sized 
network when evaluated with an ``exact method" such as Reluplex can take hours to compute. 
However, this computational speed advantage comes at the cost of over-estimation of the sensitivity measure, as both the linear relaxation 
in Eq.~\ref{eq:relaxed} and the evaluation of the dual objective at a
feasible point adds to the conservatism of the (box) over-approximation.  

\subsection{Computing Sensitivity Using Reluplex} 
\label{subsec:reluplex} 

\begin{figure}[h]
\centering
\includegraphics[width=0.35\textwidth]{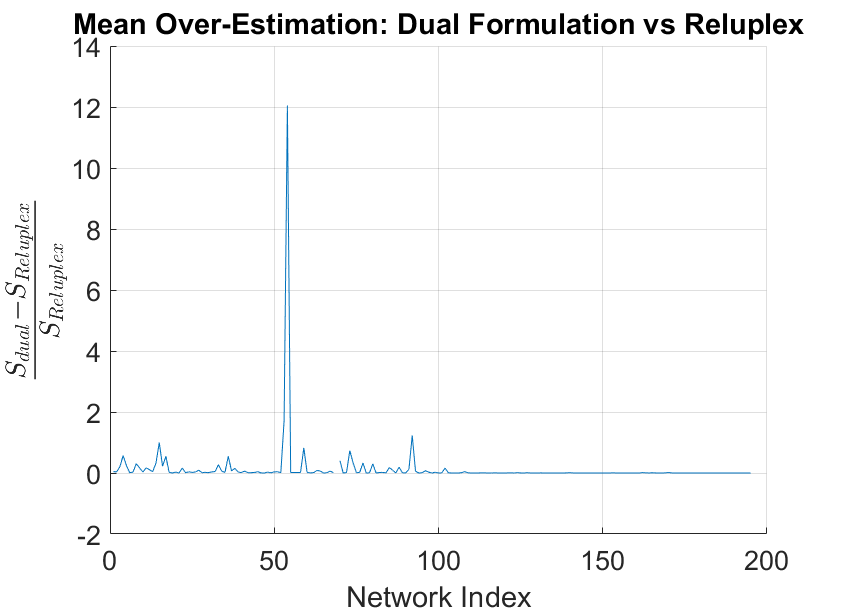}
\caption{Relative difference between sensitivity estimation via the Dual Formulation and Reluplex.} 
\label{fig:relu_vs_lp} 
\end{figure}

As mentioned in Sec.~\ref{subsec:dual}, several 
``exact" techniques benchmarked in~\cite{bunel2017piecewise} can be used to compute 
the volume of the tight over-approximation. 
In the experiments of Sec.~\ref{sec:energy}, we used Reluplex as the solver.  
To compute the volume of the tight (box) over-approximation, we queried Reluplex
the following satisfiability problem: 
for a given data input $x$ and perturbation bound $\epsilon$, 
does there exist a solution $y$ that satisfies the set of constraints
\begin{equation} 
y[i]<=a[i] \wedge y[i]>=b[i],i=1,\ldots,m 
\end{equation}
and Eqs.~\ref{eq:box},~\ref{eq:linear} and~\ref{eq:pwlinear}
in which $a[i]\in \R$ and $b[i] \in \R$ are respectively the candidate min and max of $y[i]$. 
If Reluplex
returns a negative answer, then $a[i]$ and 
$b[i]$ are valid lower and upper bounds to $y[i]$.  
Reluplex is queried repeatedly with a 
sequence of min and max candidates $\lc a[i]_j \rc$ and $\lc b[i]_j\rc$  
until a numerical tolerance is reached.     
We implemented
a simple bisection algorithm to generate a sequence of min and max candidates 
$\lc a[i]_j\rc$ and $\lc b[i]_j\rc$, and a translational tool-chain to repeatedly query Reluplex.  Fig.~\ref{fig:relu_vs_lp} plots 
$\frac{S_{dual}-S_{Reluplex}}{S_{Reluplex}}$ for close to 200 networks (each corresponding to different landscape minima) in which
$S_{dual}$ is the sensitivity computed using the dual formulation and $S_{Reluplex}$ is 
the sensitivity computed by Reluplex on the same network 
with the same input perturbation bound.  
The large peak around network \#50 shows that the dual of LP technique 
can occasionally over-estimate the sensitivity  of a network by a large amount when compared against 
Reluplex, but the plot also shows that 
for the majority of the networks the over-estimation remains relatively insignificant. 

%% file: 03energy.tex
%
In this section, we investigate the quality of the different minima in the optimization landscape of neural networks by leveraging a combination of energy landscape approaches developed by chemical physicists, and the formal verification tool Reluplex developed by the computer scientists. 

\subsection{Computing the Minima}  
\label{subsec:continuation} 
To find multiple minima, we use methods developed in computational chemistry to 
explore energy landscapes of atomic and molecular clusters, as described in~\cite{Wales03}.  
First, using different quasi-Newton methods such as limited memory Broyden-Fletcher-Goldfarb-Shanno (L-BFGS) 
and basin-hopping methods starting from different random initial guesses, we find multiple local minima. 
Next, an eigen-vector following method is used to
find multiple saddle points of index 1 again by feeding different random initial guesses and 
following the search in the restricted directions where the Hessian matrix has exactly one negative eigenvalue. 
Then, from each saddle point of index 1, we compute two steepest descent paths in two different directions 
to connect the corresponding pair of minima. 
While following this process of connecting pairs of minima, whenever any of the two end-points of this path 
is not in the existing database of the minima, it gets added to the database.
This process is iterated 
a few times to obtain multiple minima, while retaining only energetically distinct minima (minima which are at 
different geometrical locations but have the same energy value are discrete-symmetrically 
related with each other and can be obtained via a discrete transformation of weights).
To run these computations, we use a wrapper around Python Energy Landscape Explorer (PELE) \cite{pele}, which performs energy landscape related computations (i.e., all the aforementioned steps) for any given unconstrained multivariate cost function with continuous variables, as well as Theano \cite{2016arXiv160502688short}, which computes the cost function values for the given neural network architecture. To circumvent the discontinuities associated with the ReLU activation functions, we take a constant value whenever the Hessian is singular by using Theano's built-in gradients libraries. For the Iris data, we fix the $l_2$ regularization parameter to 0.001. Since this is a stochastic method, there is no guarantee that all minima will be found. We only sample the minima to obtain a qualitative picture. 

\subsection{Iris Results and Discussion} 
The network architecture selected to tackle classification on the Iris dataset consists of 3 fully-connected hidden layers of 25 neurons each. We used the energy landscape methods described in Sec.~\ref{subsec:continuation} 
to obtain 763 Iris minima of varying energy values. The computation took a total of 20 hours on a standard laptop with a single CPU and 8GB of memory. 

\begin{figure} 
\centering
\includegraphics[width=0.35\textwidth]{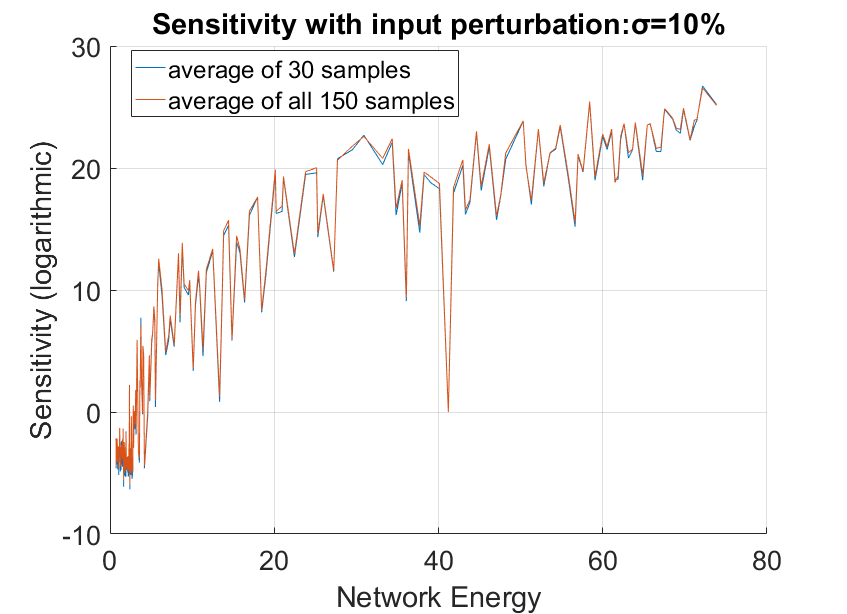}
\caption{Network sensitivity vs. network energy for Iris landscape} 
\label{fig:sense_vs_energy2} 
\end{figure}

We measure the sensitivity of each of the networks corresponding to these minima by assuming a vector of perturbation bounds $\epsilon_v=\sigma \hat{d}$ where $\sigma=0.05, 0.1, 0.2$, where $\hat{d}$ is a vector containing the ranges of values of features of the entire Iris dataset; 
that is, the perturbation bound for each input dimension is proportional to the 
spread of the input data along that dimension.  
Given a fixed $\epsilon_v$ and an input data point $x$, 
we queried Reluplex iteratively, as described in Sec.~\ref{subsec:reluplex}, 
until convergence to the min and max of each output dimension $y[i], i=1,2,3$ is achieved up to some numerical tolerance. 
The sensitivity of the network output is plotted as a function of the energy value of the landscape minima for 
$\sigma=10\%$ in Fig.~\ref{fig:sense_vs_energy2}. 
It can be seen that there is a tendency for networks corresponding to lower energy minima to have lower sensitivity and hence 
possibly exhibit more robustness to norm-bounded input perturbations. The plot illustrates the 
behavior of the average sensitivity for 30 samples and for all samples (150) of the Iris dataset. 

%% file: 05training.tex
\begin{figure}[h]
\begin{tabular}{cc}  
	\includegraphics[width=.23\textwidth]{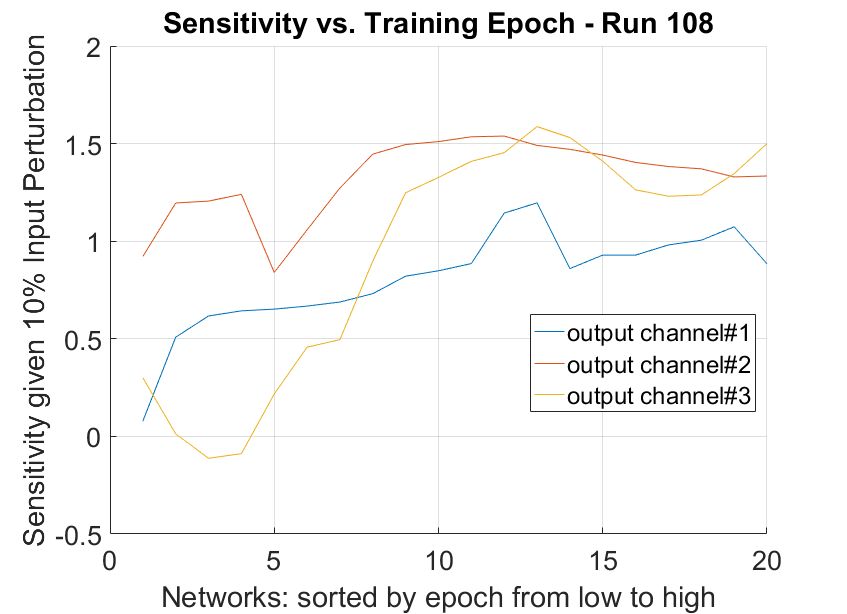} & 
	\includegraphics[width=.23\textwidth]{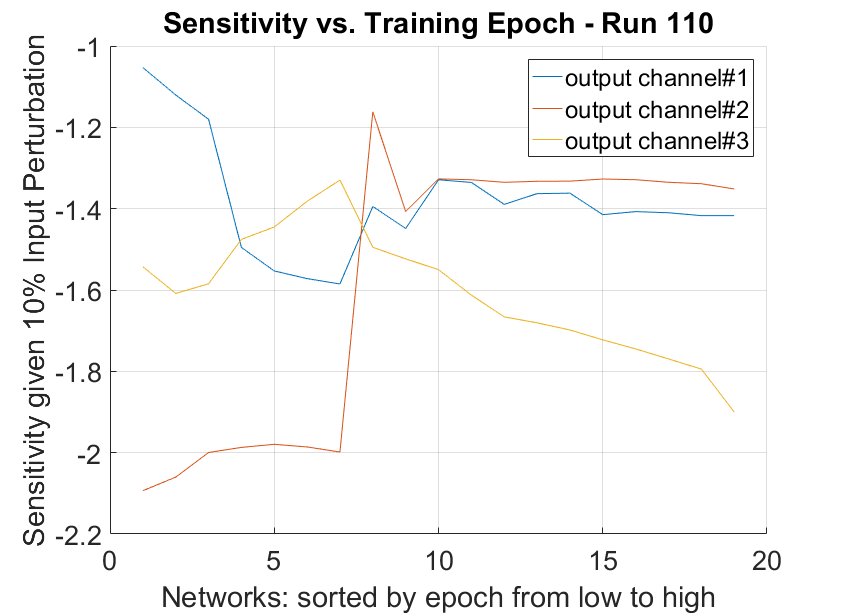} \\
	\includegraphics[width=.23\textwidth]{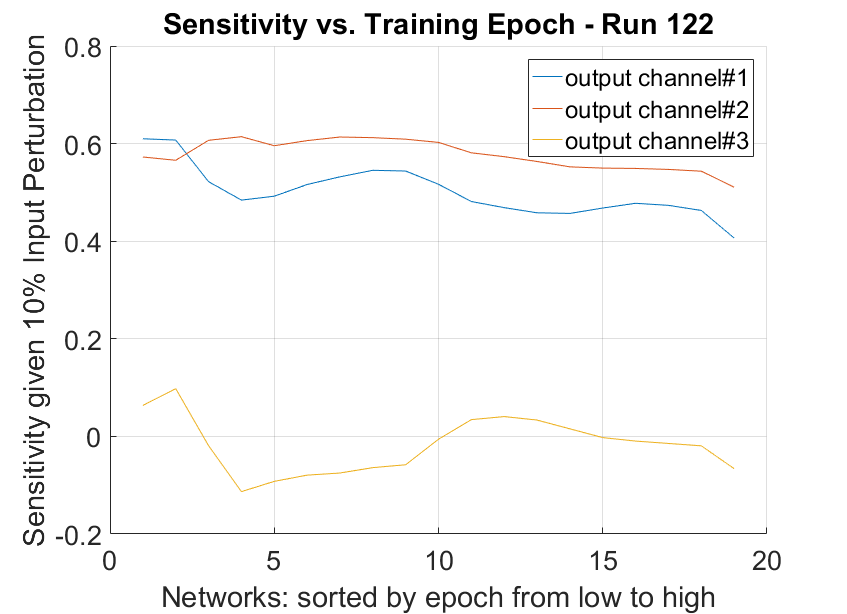} &
	\includegraphics[width=.23\textwidth]{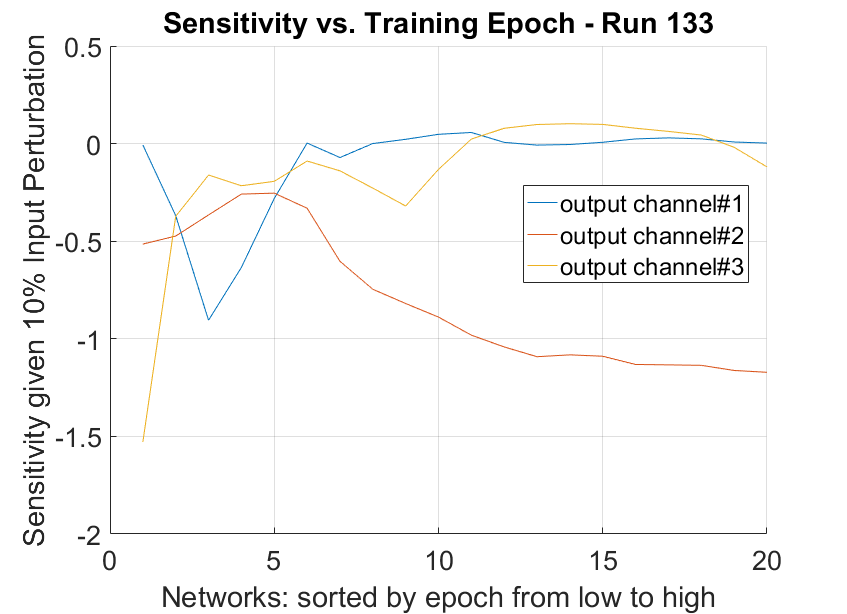}
\end{tabular} 
\caption{Samples of training runs with undesired evolution of network sensitivity.} 
\label{fig:ills} 
\end{figure}

Having shown empirically that low-lying landscape minima lead to networks that 
are less sensitive than networks with higher energy, we now describe a new training method that encourages
convergence to networks with reduced sensivitiy. 
First, we point out that the traditional task-centric loss function (e.g., cross-entropy) plus an $l_{2}$-regularization term does not necessarilty guarantee convergence to network with reduced sensitivity. Fig.~\ref{fig:ills} shows example training runs done on the Iris dataset.  The $y$-axis is the sensitivity of the network, and $x$-axis is the training epoch. It can be seen that the change in sensivitiy is irregular and without any clear pattern or trend.  

\begin{figure}[h]
\includegraphics[width=0.5\textwidth]{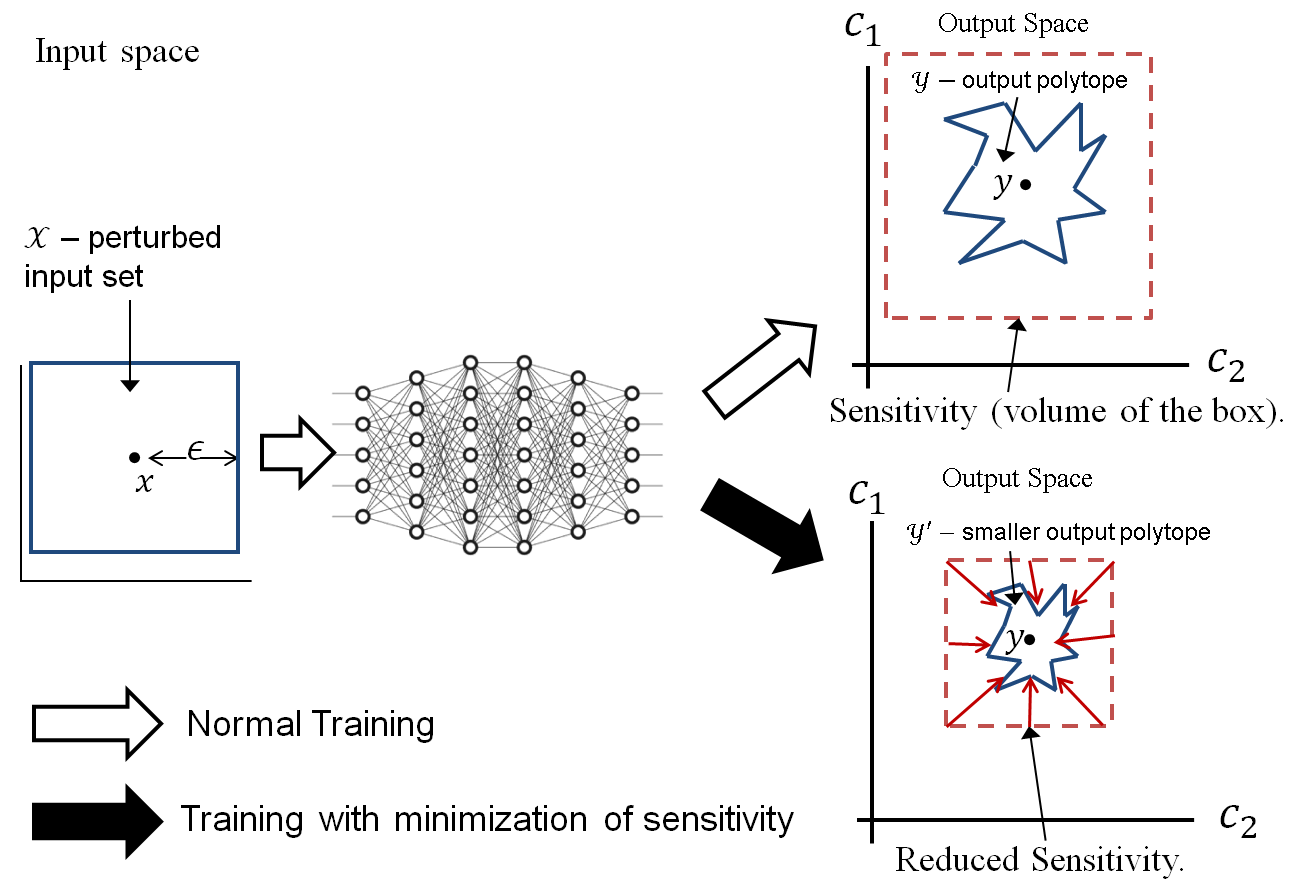}
\caption{Proposed training method to reduce network sensitivity.} 
\label{fig:training} 
\end{figure}

The proposed learning framework, which aims to minimize the sensitivity of its output networks, is illustrated in Fig.~\ref{fig:training}.
The main idea of the proposed framework is to augment the usual loss function $L$ with an additional weighted term $\lambda S(x_i,\epsilon,\nu;W)$ in which the multiplier $\lambda>0$ is a new hyperparameter, and $S(x_i,\epsilon,\nu;W)$ is the sensitivity function from~\ref{eq:sense}. Let $L(x_i,y_i;W)$ be the usual loss function with regularization incorporated. Given training sets of size $M$, the optimization task is to find the following minimizer: 
\begin{equation} 
\begin{aligned} 
	W^{*} = \underset{W}{\operatorname{argmin}} & \frac{1}{M} \sum_{i=1}^{M}  L(x_i,y_i;W) + \lambda S(x_i,\epsilon,\nu;W) 
\end{aligned} 
	\label{eq:minimizer} 
\end{equation} 
To evaluate the sensitivity function $S$ during training, any 
feasible values for the dual variables $\nu$ can be used.  The optimality of $\nu$ 
affects how conservative the estimated sensitivity will be. 
For the experiments in the next section, we evaluate $S$ using Algorithm 1 from~\cite{Kolter2017}.

%% file: 06results.tex
In this section, we present two sets of experimental results comparing the 
performance of our training technique across several metrics 
against other state-of-art techniques in the literature. 
The standard handwritten digits data-set MNIST is used in all experiments.  
The baseline training for all comparisons consists of a cross-entropy loss function with $l_2$ regularization.  

In the first set of experiments, we compare our our sensitivity minimization (SM) algorithm against two other techniques which, like the proposed approach, minimize some measure of sensitivity.  The first technique~\cite{DBLP:journals/corr/abs-1803-08680} (jacobM) performs regularization on the Frobeonius norm of the local Jacobian matrix of the network's classification function, while the second method~\cite{Yoshida2017SpectralNR} (specM) implements a regularization scheme on the spectral norm of the network weights.
The results from the first set of experiments are discussed in Sec.~\ref{subsec:senses}. 

The second set of experiments aims at demonstrating that some improvement in performance against adversarial attacks results as a side consequence of sensitivity minimization. To that end, we compare SM against the described baseline approach as well as the algorithm from~\cite{Kolter2017} (K\&W), which maximizes an adversarial robustness measure explicitly. The results of these experiments are discussed in Sec.~\ref{subsec:kw}.

In both experiments, the weight on our sensitivity loss was chosen to be $\lambda=1\times 10^{-6}$.      
This value was empirically optimized based on multiple training runs and provides a 
good balance between the classification performance and sensitivity of the resulting network. 
A perturbation bound as described in Sec.~\ref{sec:energy} with $\sigma = 0.05$ was chosen for all evaluations of sensitivity and 
adversarial robustness.  
One training instance with identical sets of mini-batches is executed for each method being tested, and each resulting trained network is evaluated across all metrics under consideration. Performance is measured on both the training and test data. The standard MNIST training and test sets are used throughout. 

\subsection{Experiments on Sensitivity-Based Optimization}
\label{subsec:senses} 
The network architecture consists of two convolutional layers and one fully-connected layer. We use the Adam Optimizer.  As stated, the competing training techniques are applied on identical sets of mini-batches. The set of training techniques include
SM, specM, jacobM, as well as the baseline. The metrics used to evaluate performance are cross-entropy loss (CE), classification error (ERR), our own sensitivity measure from Eq.~\ref{eq:minimizer} (SENSE), the Jacobian loss from~\cite{DBLP:journals/corr/abs-1803-08680} (JACOB), the Spectral loss from~\cite{Yoshida2017SpectralNR} (SPECTRAL), and K\&W's~\cite{Kolter2017} adversarial loss (ADV\_LOSS) and (ADV\_ERR) (i.e., the percentage of the input data for which the application of bounded perturbations could lead to adversarial examples).  

The metric values (after convergence) for each of the techniques are listed in Table~\ref{tab:results02}. Top performers are highlighted (dark green), as well as second-best performers that are closer to the top performer than to the bottom performer (lime green). The plots of the evolutions of the metrics through the training process are provided in the appendix. 
\begin{table}
\caption{Performance results on MNIST test data (sensitivity metrics - smaller is better).} 
\resizebox{0.5\textwidth}{!}{%
	\begin{tabular}{lcccc} 
		Method/Metric   &   CE     & ERR      & ADV\_LOSS  &  ADV\_ERR  \\
		Base & \textcolor{ForestGreen}{$\mathbf{0.033}$} & \textcolor{ForestGreen}{$\mathbf{0.92}$\% }& $12.12$    & $96.55$\%  \\
		specM & \textcolor{LimeGreen}{$\mathbf{0.034}$} & \textcolor{LimeGreen}{$\mathbf{0.95}$\% } & $10.31$   & $97.53$\%  \\
		jacobM & $0.155$ &  $2.74$\% &\textcolor{LimeGreen}{ $\mathbf{2.34}$   } & $71.73$\%  \\
		SM    &  $0.079 $ & $2.26$\% &\textcolor{ForestGreen}{ $\mathbf{0.41}$ }   & \textcolor{ForestGreen}{$\mathbf{12.62}$\% }  \\
		\\
		Method/Metric &  SENSE                  & JACOB     & SPECTRAL &  \\
		Base. & $4.58\times 10^{12}$ & $17.47$ & $7.19$ &   \\
		specM &  $71.1\times 10^{12}$ &  $12.49$ &\textcolor{ForestGreen}{ $\mathbf{4.17}$ }&   \\
		jacobM & \textcolor{LimeGreen}{$\mathbf{44.5\times 10^{6}}$ }& \textcolor{ForestGreen}{ $\mathbf{2.75}$ }&\textcolor{LimeGreen}{$\mathbf{5.07}$ }&  \\
		SM &  \textcolor{ForestGreen}{ $\mathbf{5.53\times 10^{3}} $ }& \textcolor{LimeGreen} { $\mathbf{7.75}$ }& $6.83$ &  
\end{tabular} }
\label{tab:results02} 
\end{table} 

When compared to the baseline, the results show that the spectral regularization technique has very small impact on the adversarial loss and little to no effect on the adversarial error. Jacobian loss regularization performed better than spectral regularization in terms of those robustness measures; however it still lags behind our sensitivity minimization technique in terms of guaranteeing robustness to adversarial samples. Any increase in robustness once again comes at the cost of classification performance as both jacobM and SM are behind specM and the baseline in that regard.  Our method handily outperforms all competing methods on the adversarial-related metrics.

In terms of the sensitivity measures (i.e., SENSE, JACOB and SPECTRAL), 
each training method is the best at minimizing its own metric.  Regarding the cross-regularizations performance, jacobM outperformed specM by many orders of magnitude at minimizing our sensitivity measure.  It is also better than our technique at minimizing the spectral norm. Our technique did perform better than specM at minimizing the Jacobian loss.  

\subsection{Experiments on Adversarial-Margin-Based Optimization}  
\label{subsec:kw} 
For the comparison between baseline, SM and K\&W, we use a network architecture consisting of one convolutional layer and one fully-connected layer. For this experiment, we used a vanilla stochastic gradient descent (SGD) optimizer without additional 
heuristics. We compare performance on CE, ERR, SENSE, ADV\_LOSS and ADV\_ERR. 


The final performance numbers for the training and test data are listed in Tables~\ref{tab:results01} and~\ref{tab:results01_}. For more detailed comparisons of how the performance of the techniques on a particular metric evolved through 
the course of the training, see figures in the appendix.
\begin{table}
	\caption{Performance results on MNIST training data (adversarial metrics - smaller is better)} 
\resizebox{0.5\textwidth}{!}{%
	\begin{tabular}{lcccccc} 
T/M        &  CE     & ERR &     SENSE   &    ADV\_LOSS  &  ADV\_ERR  \\
Base. &  \textcolor{ForestGreen}{$\mathbf{0.071}$} &  \textcolor{ForestGreen}{$\mathbf{1.94}$\%} & $1.28 \times 10^{12}$ & $6.25$ & $91.98$\%  \\
		K\&W &  $0.285$ &  $8.02$\% & \textcolor{LimeGreen}{$\mathbf{2.0 \times 10^{5}}$} &  \textcolor{ForestGreen}{$\mathbf{0.55}$ }&  \textcolor{ForestGreen}{$\mathbf{16.02}$\% } \\
	SM &  \textcolor{LimeGreen}{$\mathbf{0.138}$ }&  \textcolor{LimeGreen}{$\mathbf{3.56}$\% } & \textcolor{ForestGreen}{$\mathbf{8.76\times10^{3}}$ }&  \textcolor{LimeGreen}{$\mathbf{0.85}$ }& \textcolor{LimeGreen}{$\mathbf{26.36}$\% }
\end{tabular} }
\label{tab:results01} 
\end{table} 
\begin{table}
\caption{Performance results on MNIST test data (adversarial metrics - smaller is better)} 
\resizebox{0.5\textwidth}{!}{%
	\begin{tabular}{lcccccc} 
		T/M   &   CE    & ERR      &  SENSE   &    ADV\_LOSS  &  ADV\_ERR  \\
		Base &  \textcolor{ForestGreen}{$\mathbf{0.082}$ }&  \textcolor{ForestGreen}{$\mathbf{2.38}$\% }& $1.27 \times 10^{12}$& $6.11$ & $91.23$\%  \\
		K\&W &  $0.262$ &  $7.3$\% &  \textcolor{LimeGreen}{$\mathbf{2.12 \times 10^{5}}$} &  \textcolor{ForestGreen}{$\mathbf{0.51}$ }&  \textcolor{ForestGreen}{$\mathbf{14.54}$\%} \\
		SM &  \textcolor{LimeGreen}{$\mathbf{0.125}$ }&  \textcolor{LimeGreen}{$\mathbf{3.25}$\% } &  \textcolor{ForestGreen}{$\mathbf{8.65 \times10^{3}}$ }&  \textcolor{LimeGreen}{$\mathbf{0.79}$} &  \textcolor{LimeGreen}{$\mathbf{24.5}$\%} 
\end{tabular} }
\label{tab:results01_} 
\end{table} 

Our technique outperforms both the baseline and K\&W on the sensitivity measure. 
In terms of the adversarial error, our technique was outperformed by K\&W; however, when compared against the 
baseline, both methods increase the robustness of the network by significant margins. 
Clearly, there is a correlation between the lower sensitivities and lower adversarial loss/error. While our technique does not minimize the adversarial error directly, 
the empirical outcome indicated in Table~\ref{tab:results01} 
seems to indicate that encouraging small sensitivity has an effect on that measure. 
Complementarily, the method from~\cite{Kolter2017} directly minimizes the adversarial error by maximizing the adversarial margin; however, as shown in Table~\ref{tab:results01}, it appears to also have significant effect towards minimizing the 
sensitivity of the network when compared against the baseline.

In terms of classification error and cross-entropy loss, the performance of our training method exceeds that of K\&W. We believe this difference in classification performance is due to overfitting in K\&W taking place.  This could be due to the fact that K\&W learns boundaries based on worst-case adversarial examples, which may lead to unnecessarily uneven boundaries, often associated with overfitting; in contrast, our training method reduces the size of the output reachable set uniformly, and not based on any individual sample.  The significant gains in classification performance and sensitivity on both training and test sets of our method relative to K\&W, more than make up for the comparatively slight loss in adversarial performance. 



%% file: 07conclu.tex
In this paper, we studied the relationship between the energy value of optimization landscape minima and the sensitivity and adversarial robustness of the resulting networks. Using formal verification and energy landscape techniques, we empirically showed that there exists a correlation between the sensitivity of a network and the energy value of its corresponding landscape minimum: the lower the energy of the minimum, the less sensitive the network is to perturbations in the input. We also studied the relationship between the sensitivity of a neural network and its adversarial robustness, and show that less sensitive networks tend to be more robust. Furthermore, we found that there is a consistent trade-off between adversarial and sensitivity properties of a network and its classification performance.

Leveraging those experimental findings, we introduced a novel learning framework aimed at optimizing the sensitivity of a network, as measured by the aggregate volume of an over-approximation of the reachable set of network outputs under bounded additive perturbations.  We showed experimentally that, as expected, the proposed sensitivity-based learning approach had positive impact on the adversarial robustness of the resulting networks.  Overall, we found that the proposed method compares favorably against state-of-the-art sensitivity-based learning frameworks with regards to adversarial robustness, while at the same time remaining competitive across miscellaneous sensitivity metrics.

An additional set of experiments comparing the performance of the proposed network and an adversarially-tuned state-of-the-art approach showed that application of our method results in a reasonable trade-off between sensitivity, adversarial robustness and classification performance.  While the proposed method slightly lagged behind in terms of the computed adversarial metrics, we believe the observed sensitivity gains and ameliorated losses in classification performance represent a more than sensible compromise which may be preferred in certain applications.



%% file: appendix02.tex
\label{subsec:append02} 
Note that the shifts in behavior observed in the plots 
describing the evolution of jacobM~\cite{DBLP:journals/corr/abs-1803-08680} 
are due to the fact that the method consists of three separate training stages: 
between epochs 1 and 41, it is equivalent to the baseline method; between epochs 41 and 59, it incorporates dropout on top of the baseline approach; lastly, from epoch 61 on, it enforces the Jacobian loss term. 
\begin{figure}[H]
    \centering
    \begin{subfigure}[b]{0.475\textwidth}
        \includegraphics[width=1.0\textwidth]{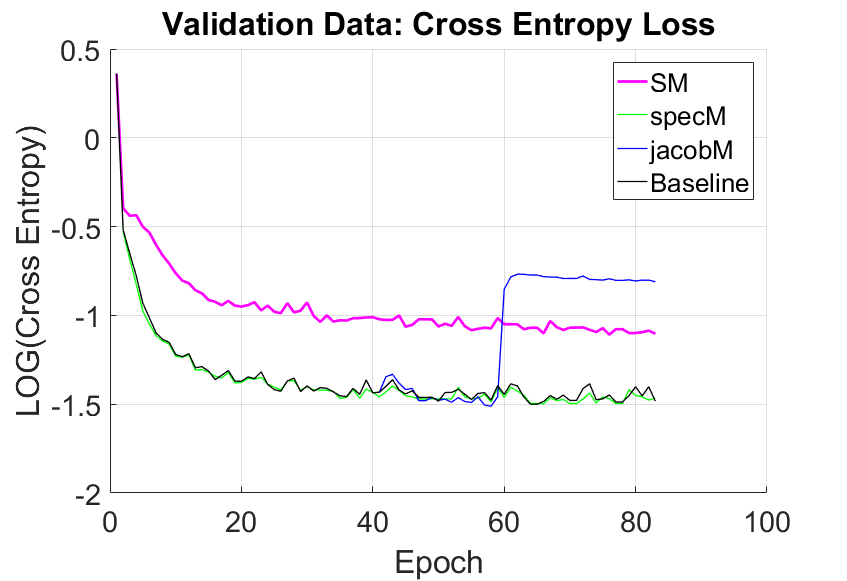}
        \caption{Cross-Entropy Loss on Validation Data}
        \label{fig:ce_loss_test_sense}
    \end{subfigure}
    \begin{subfigure}[b]{0.475\textwidth}
        \includegraphics[width=1.0\textwidth]{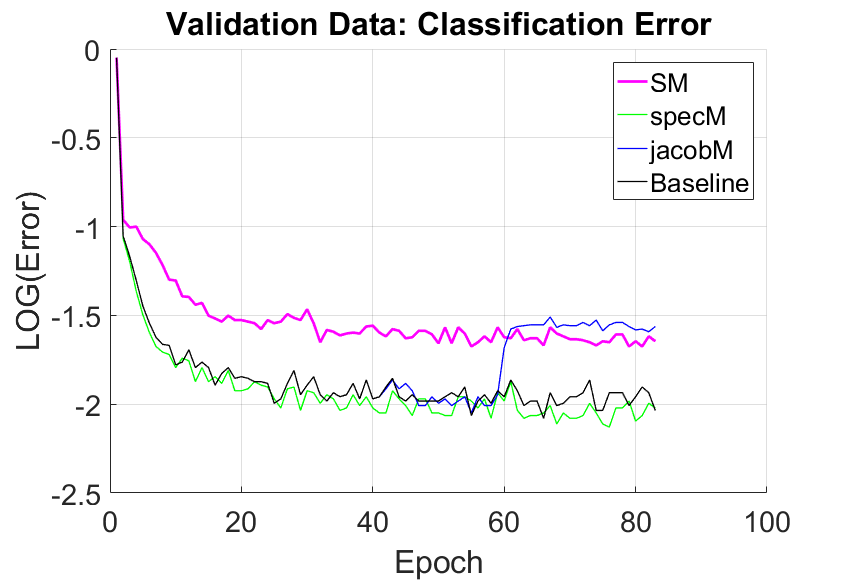}
        \caption{Classification Error on Validation Data}
        \label{fig:class_error_test_sense}
    \end{subfigure}
    \caption{Performance Comparison - Performance Metrics}
\label{fig:runs02_ce}
\end{figure}

\begin{figure}[H]
\centering
        \includegraphics[width=0.6\textwidth]{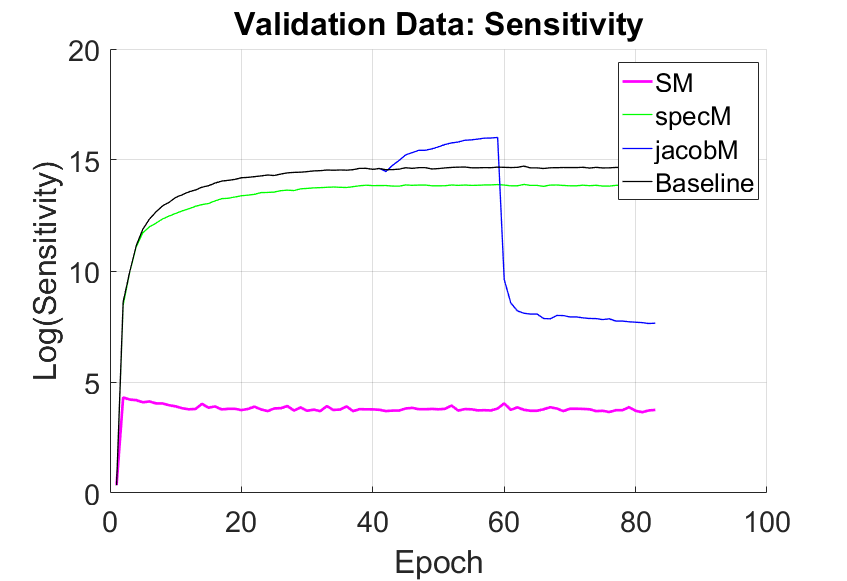}
        \caption{Sensitivity on Validation Data}
        \label{fig:sense_test_sense}
\end{figure} 

\begin{figure}[H]
    \begin{subfigure}[b]{0.475\textwidth}
        \includegraphics[width=1.0\textwidth]{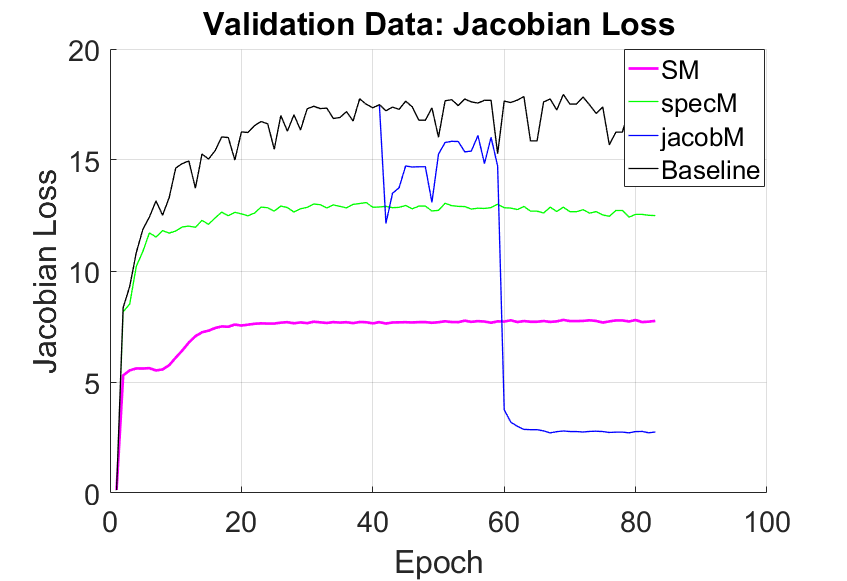}
        \caption{Jacobian Loss on Validation Data}
        \label{fig:jacob_loss_test_sense}
    \end{subfigure}
    \begin{subfigure}[b]{0.475\textwidth}
        \includegraphics[width=1.0\textwidth]{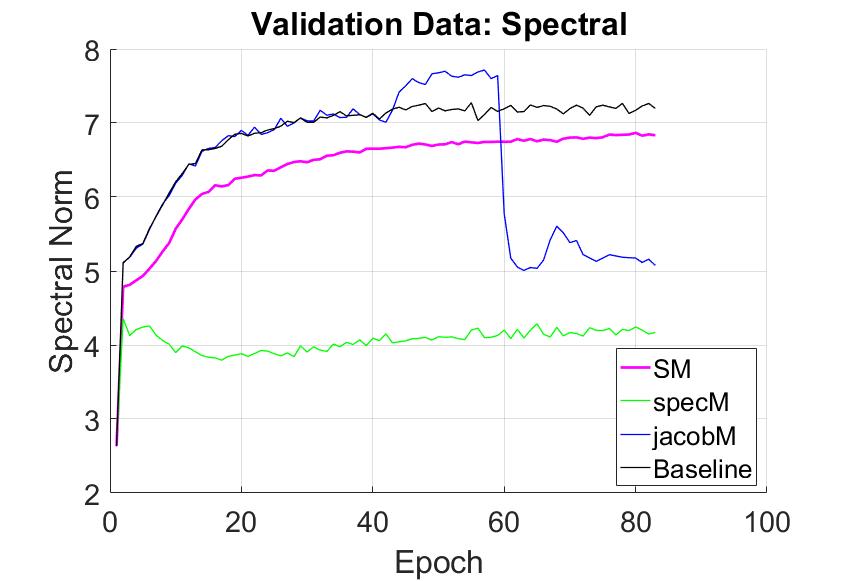}
        \label{fig:spectral_loss_test_sense}
	 \caption{Spectral Loss on Validation Data}
    \end{subfigure}
    \caption{Performance Comparison - Other Sensitivity Metrics}
\label{fig:runs02_sense2}
\end{figure}

\begin{figure}[H]
    \begin{subfigure}[b]{0.475\textwidth}
        \includegraphics[width=1.0\textwidth]{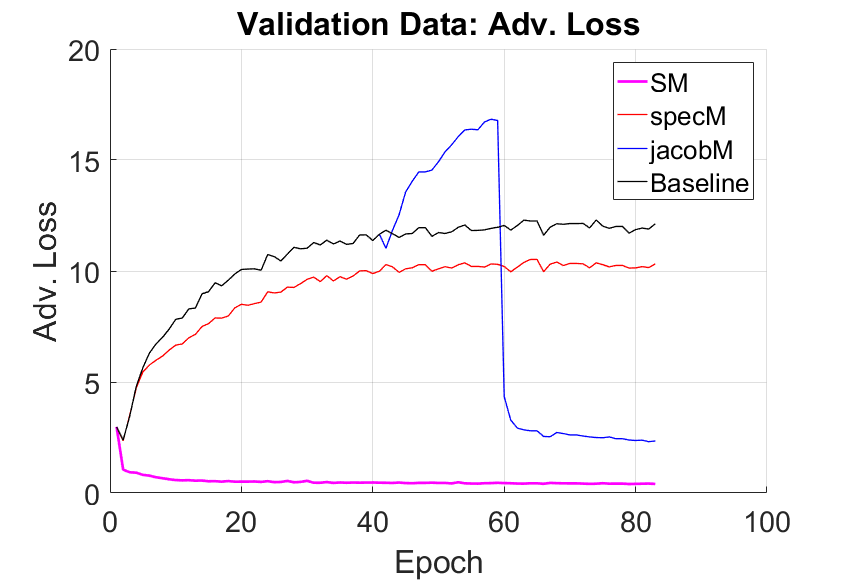}
        \caption{Adversarial Loss on Validation Data}
        \label{fig:adv_loss_test_sense}
    \end{subfigure}
    \begin{subfigure}[b]{0.475\textwidth}
        \includegraphics[width=1.0\textwidth]{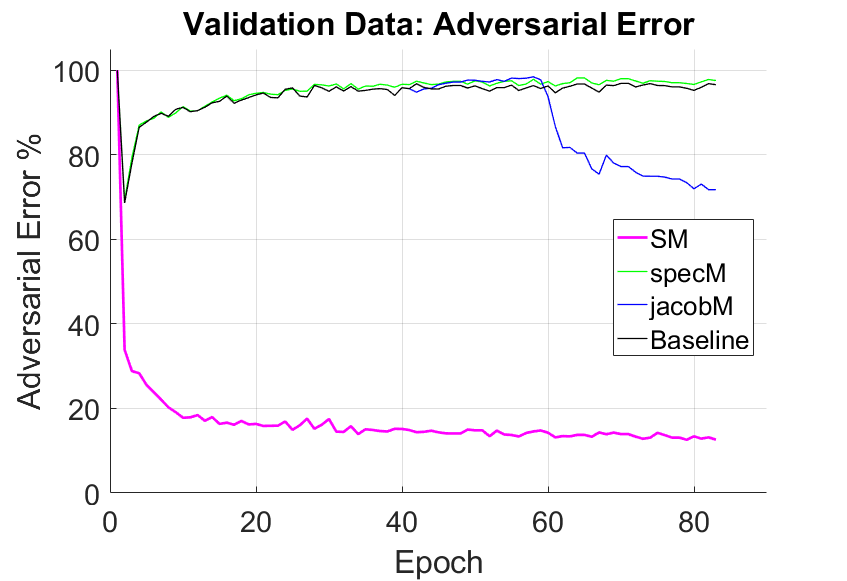}
        \caption{Adversarial Error on Validation Data}
        \label{fig:adv_error_test_sense}
    \end{subfigure}
    \caption{Performance Comparison - Adversarial Metrics} 
\label{fig:runs02_adv}
\end{figure}

%% file: appendix01.tex
\label{subsec:append01} 
Note that the granularity of all training data plots is 250 mini-batches.  
\begin{figure}[H]
    \centering
    \begin{subfigure}[b]{0.465\textwidth}
        \includegraphics[width=1.0\textwidth]{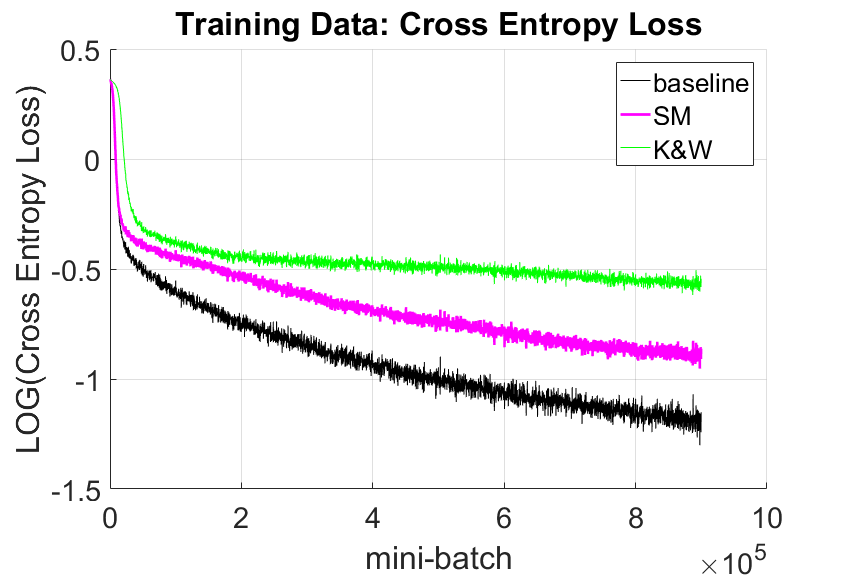}
        \caption{Cross-Entropy Loss on Training Data}
        \label{fig:ce_loss_train_kw}
    \end{subfigure}
    \begin{subfigure}[b]{0.465\textwidth}
        \includegraphics[width=1.0\textwidth]{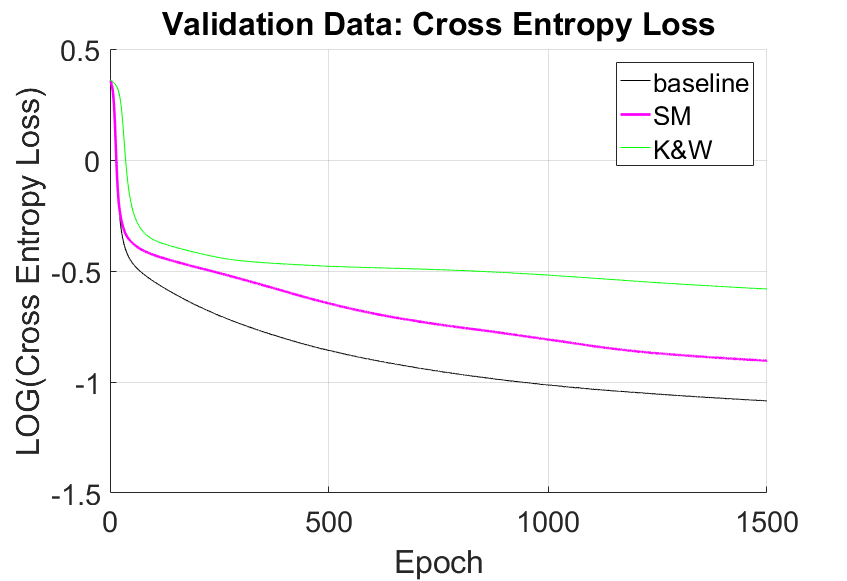}
        \caption{Cross-Entropy Loss on Validation Data}
        \label{fig:ce_loss_test_kw}
    \end{subfigure}
    \caption{Performance Comparison - Cross-Entropy Loss}
\label{fig:runs01_ce}
\end{figure}

\begin{figure}[H]
    \begin{subfigure}[b]{0.465\textwidth}
        \includegraphics[width=1.0\textwidth]{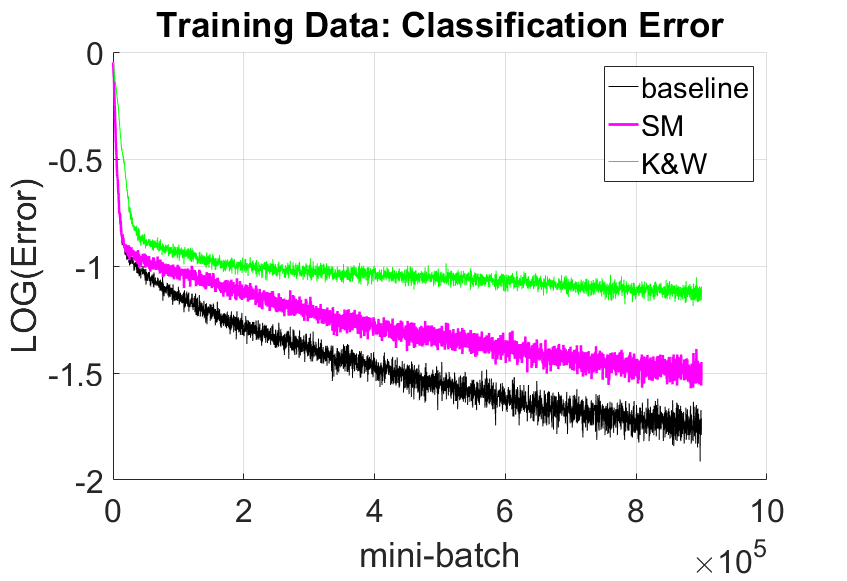}
        \caption{Classification Error on Training Data}
        \label{fig:class_error_train_kw}
    \end{subfigure}
    \begin{subfigure}[b]{0.465\textwidth}
        \includegraphics[width=1.0\textwidth]{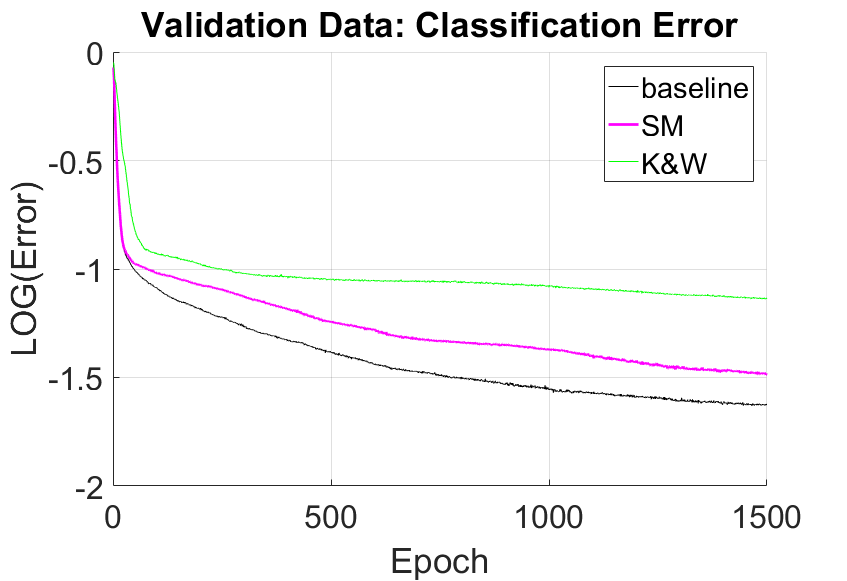}
        \caption{Classification Error on Validation Data}	
        \label{fig:class_error_test_kw}
    \end{subfigure}
    \caption{Performance Comparison - Classification Error Rate}
\label{fig:runs01_class}
\end{figure}

\begin{figure}[H]
    \begin{subfigure}[b]{0.465\textwidth}
        \includegraphics[width=1.0\textwidth]{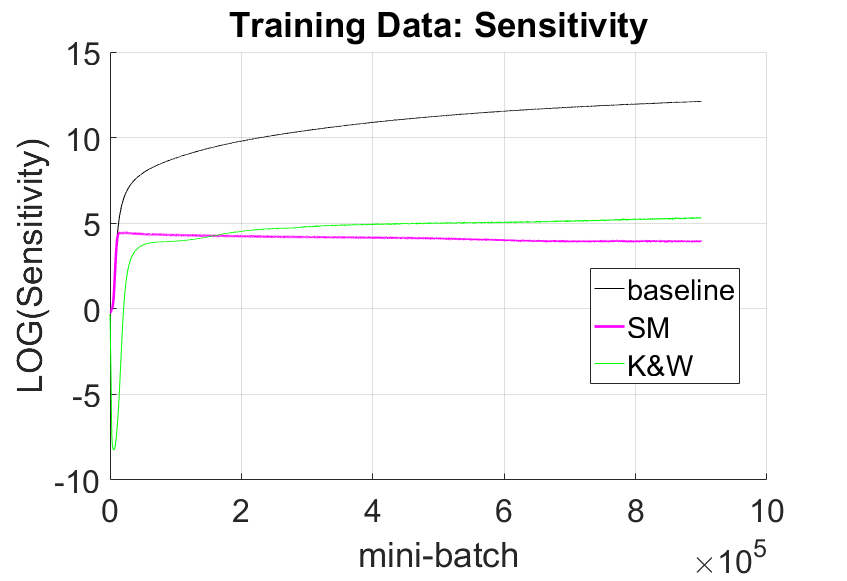}
        \caption{Sensitivity on Training Data}
        \label{fig:sense_train_kw}
    \end{subfigure}
    \begin{subfigure}[b]{0.465\textwidth}
        \includegraphics[width=1.0\textwidth]{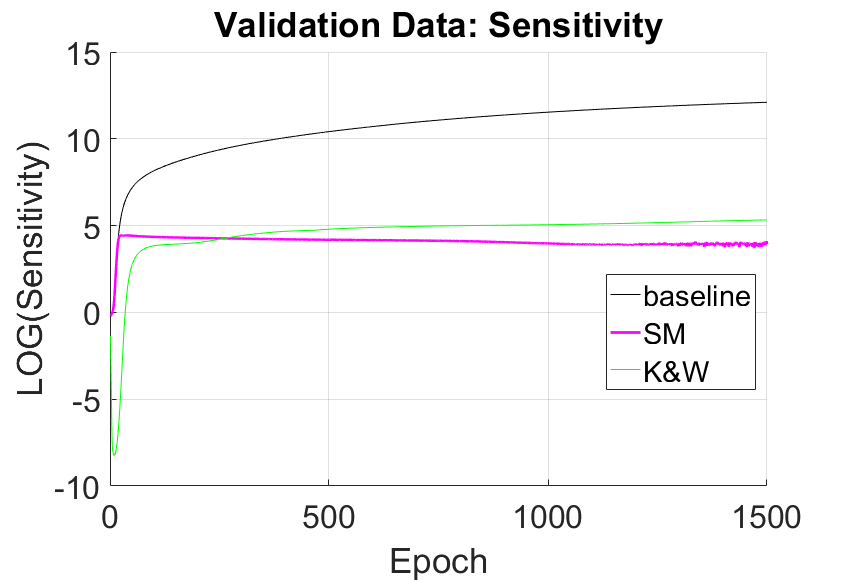}
        \caption{Sensitivity on Validation Data}
        \label{fig:sense_test_kw}
    \end{subfigure}
    \caption{Performance Comparison - Sensitivity}
\label{fig:runs01_sense}
\end{figure}

\begin{figure}[H]
    \begin{subfigure}[b]{0.465\textwidth}
        \includegraphics[width=1.0\textwidth]{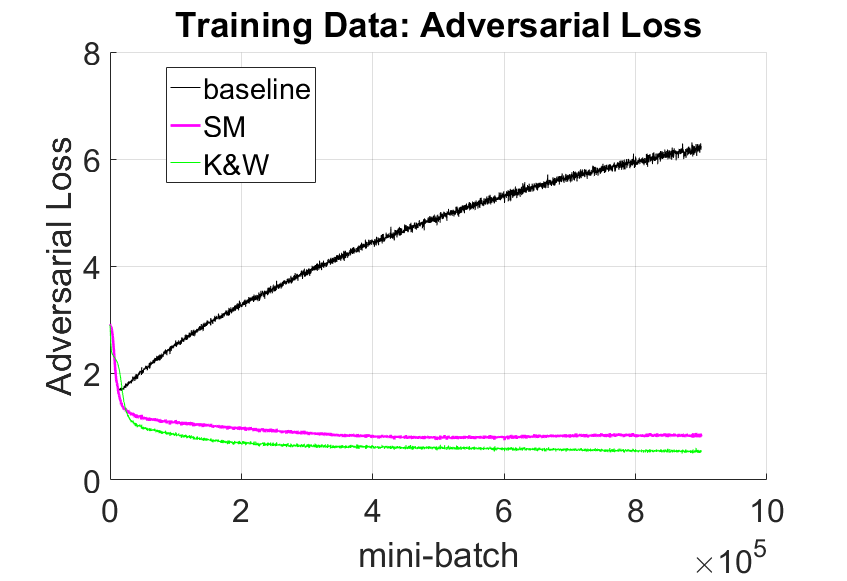}
        \caption{Adversarial Loss on Training Data}
        \label{fig:adv_loss_train_kw}
    \end{subfigure}
    \begin{subfigure}[b]{0.465\textwidth}
        \includegraphics[width=1.0\textwidth]{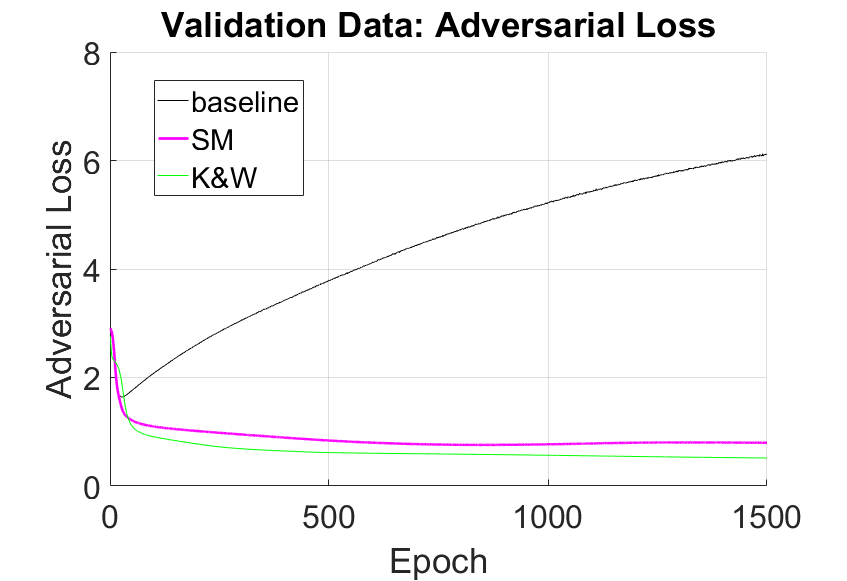}
        \caption{Adversarial Loss on Validation Data}
        \label{fig:adv_loss_test_kw}
    \end{subfigure}
    \caption{Performance Comparison - Adversarial Loss}
\label{fig:runs01_adv_loss}
\end{figure}

\begin{figure}[H]
    \begin{subfigure}[b]{0.465\textwidth}
        \includegraphics[width=1.0\textwidth]{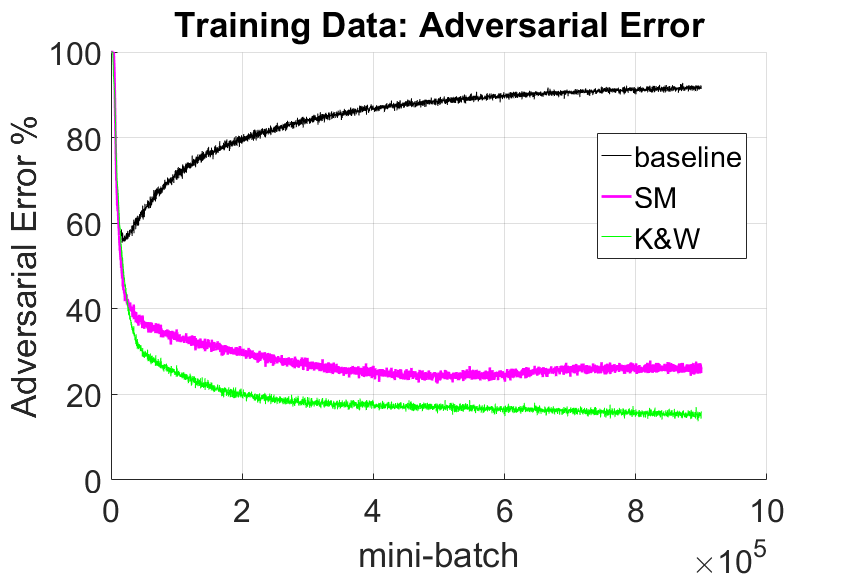}
        \caption{Adversarial Error Rate on Training Data}
        \label{fig:adv_error_train_kw}
    \end{subfigure}
    \begin{subfigure}[b]{0.465\textwidth}
        \includegraphics[width=1.0\textwidth]{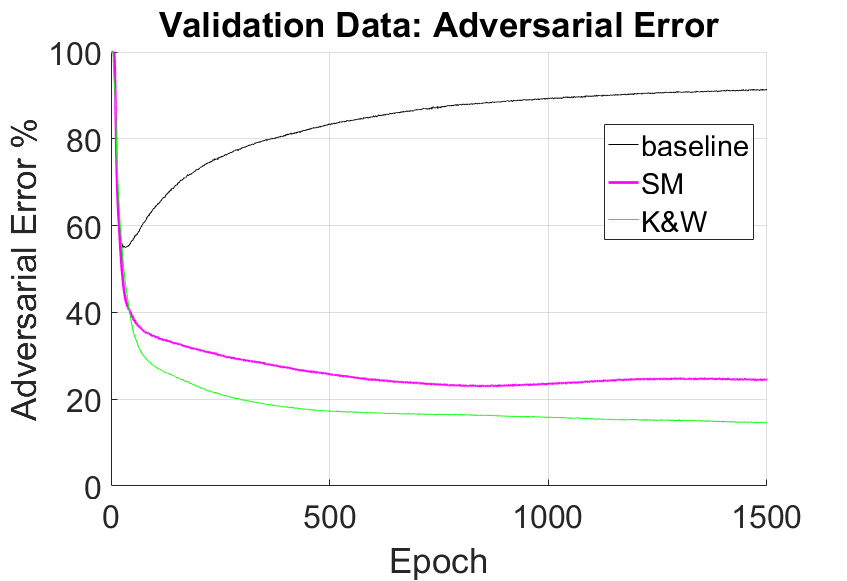}
        \caption{Adversarial Error Rate on Validation Data}
        \label{fig:adv_error_test_kw}
    \end{subfigure}
    \caption{Performance Comparison - Adversarial Error Rate}
\label{fig:runs01_adv_error}
\end{figure}